\documentclass[runningheads]{llncs}

\usepackage{eccvabbrv}

\usepackage{graphicx}
\usepackage{booktabs}

\usepackage[accsupp]{axessibility}  
\usepackage[normalem]{ulem}
\usepackage{graphicx}   
\usepackage{booktabs}   
\usepackage{multirow}   
\usepackage{amsmath}        
\usepackage{amssymb}        
\usepackage{bm}             
\DeclareUnicodeCharacter{2212}{-} 
\DeclareUnicodeCharacter{2191}{\ensuremath{\uparrow}} 
\DeclareUnicodeCharacter{2193}{\ensuremath{\downarrow}} 
\usepackage{svg}
\DeclareUnicodeCharacter{2212}{-}
\usepackage{algorithm}
\usepackage{algpseudocode}
\usepackage{placeins}
\usepackage{pifont}
\usepackage[table,xcdraw]{xcolor}
\usepackage[x11names]{xcolor}
\usepackage{makecell}
\usepackage{booktabs}    
\usepackage{array}       
\usepackage{multirow}    
\usepackage{caption}     
\usepackage{threeparttable} 
\usepackage{tabularx}       
\usepackage[table]{xcolor} 
\usepackage{wrapfig}
\usepackage{caption} 
\usepackage{adjustbox}
\usepackage{bbding}

\usepackage{hyperref}
\usepackage{orcidlink}

\begin{document}

\title{GeoAlignCLIP: Enhancing Fine-Grained Vision-Language Alignment in Remote Sensing via Multi-Granular Consistency Learning} 

\titlerunning{ }

\author{
Xiao Yang\inst{1,2} \and
Ronghao Fu\inst{1,2}\Envelope \and
Zhuoran Duan\inst{1,2} \and Zhiwen Lin\inst{1,2} \and Xueyan Liu\inst{1,2} \and Bo Yang\inst{1,2}\Envelope
}

\authorrunning{Yang et al.}
\institute{
$^1$College of Computer Science and Technology, Jilin University, Changchun 130012, China\\
$^2$Key Laboratory of Symbolic Computation and Knowledge Engineering of Ministry of Education Jilin University \\
\email{\{yangx23,duanzr24,linzw25\}@mails.jlu.edu.cn, \{furh,xueyanliu,ybo\}@jlu.edu.cn}}

\def\customsymbol#1{
    \ifcase\number\value{#1}
        \or\Envelope
    \else\@ctrerr
    \fi
}

\maketitle
\renewcommand{\footnotesize}{\fontsize{8pt}{8pt}\selectfont}
\renewcommand{\thefootnote}{\customsymbol{footnote}}
\footnotetext[1]{Corresponding author.}

\begin{abstract}
  Vision-language pretraining models have made significant progress in bridging remote sensing imagery with natural language. However, existing approaches often fail to effectively integrate multi-granular visual and textual information, relying primarily on global image-text alignment. This limitation hinders the model's ability to accurately capture fine-grained details in images, thus restricting its performance in complex, fine-grained tasks. To address this, we propose GeoAlignCLIP, a unified framework that achieves fine-grained alignment in remote sensing tasks by learning multi-granular semantic alignments and incorporating intra-modal consistency, enabling more precise visual-semantic alignment between image regions and text concepts. Additionally, we construct RSFG-100k, a fine-granular remote sensing dataset containing scene descriptions, region-level annotations, and challenging hard-negative samples, providing hierarchical supervision for model training. Extensive experiments conducted on multiple public remote-sensing benchmarks demonstrate that GeoAlignCLIP consistently outperforms existing RS-specific methods across diverse tasks, exhibiting more robust and accurate fine-grained vision-language alignment.
  \keywords{Remote Sensing \and VLM \and Fine-Grained Alignment}
\end{abstract}

\section{Introduction}
\label{sec:1_introduction}
\vspace{-0.3cm}

Vision–Language Models (VLMs) learn joint visual–semantic representations by aligning images with textual descriptions. As a representative paradigm, Contrastive Language-Image Pre-training (CLIP)~\cite{radford2021learning} has become the cornerstone of vision-language representation learning. By aligning image and text embeddings in a shared semantic space through large-scale contrastive training, CLIP has demonstrated remarkable zero-shot transfer and open-domain generalization capabilities. Building on its success, Remote Sensing (RS) community has developed a series of domain-adapted CLIP variants, which have greatly advanced multimodal understanding and interpretation of RS imagery. However, despite their impressive capabilities, these models often exhibit clear limitations in capturing fine-grained visual details and contextual semantics, particularly when recognizing objects with subtle inter-class variations or complex spatial relationships. Moreover, the absence of supervision for fine-grained textual semantics further hinders their capability to achieve precise and context-aware semantic alignment.

\begin{figure}[t]
  \centering
\includegraphics[width=0.95\linewidth]{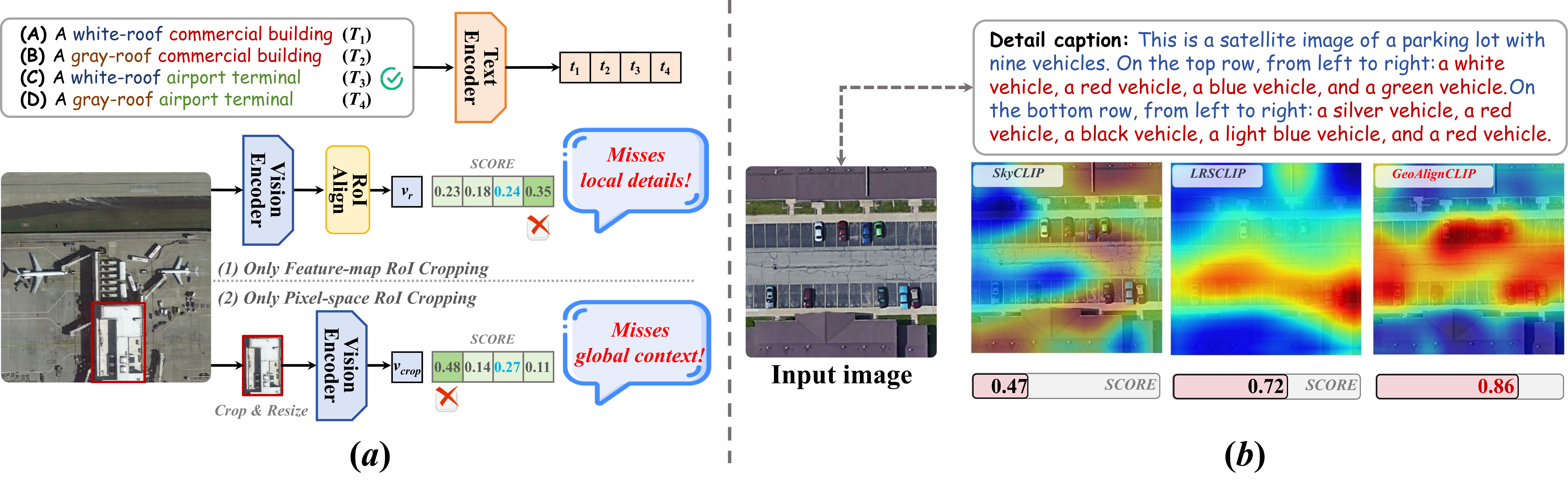}
  \vspace{-0.45cm}
  \caption{(a) Comparison of feature-map RoI cropping and pixel-space RoI cropping methods for vision-language alignment.  
  (b) Comparison of attention heatmaps for models trained with different text granularities (brief, enriched, and multi-granular) in capturing global and local semantic relationships.}
  \label{fig:intro_motivation}
  \vspace{-0.7cm}
\end{figure}

Existing RS-specific CLIPs, such as RemoteCLIP~\cite{liu2024remoteclip} and GeoCLIP~\cite{vivanco2023geoclip}, have achieved remarkable progress in adapting vision–language pre-training to RS imagery. However, most of them still suffer from performance degradation in fine-grained recognition and cross-domain alignment tasks. Specifically, as illustrated in Figure \ref{fig:intro_motivation}, current domain-adapted CLIP frameworks still encounter several key challenges: \textbf{1) Limitations in global–local visual representation.} As shown in Figure \ref{fig:intro_motivation}(a), feature-map Region of Interest (RoI) cropping enables models to roughly localize semantic regions (e.g., identifying parts of an airport terminal), but it struggles to extract discriminative features from dense visual scenes, resulting in ambiguous recognition under complex backgrounds. In contrast, pixel-space RoI cropping captures more localized details yet sacrifices global contextual awareness, leading to confusion between visually similar structures, such as commercial buildings and airport terminals. These findings reveal that effectively balancing global and local visual representations is crucial for reliable fine-grained understanding. \textbf{2) Limitations in multi-granular textual alignment.} As shown in Figure \ref{fig:intro_motivation}(b), models trained with brief text captions (e.g., SkyCLIP~\cite{wang2024skyscript}) tend to focus on coarse global semantics while neglecting inter-object relationships (e.g., vehicle distribution). Conversely, models using enriched long-text descriptions (e.g., LRSCLIP~\cite{chen2025lrsclip}) capture local semantics but lose global consistency, leading to fragmented vision-language correspondence. These observations highlight the need for multi-granular textual modeling that integrates both global context and object-level details, thereby ensuring coherent fine-grained alignment.

To overcome the limitations of existing RS-specific CLIP models in fine-grained recognition and cross-domain alignment under complex spatial layouts and high inter-class similarity, we propose \textbf{GeoAlignCLIP}, a unified vision--language framework tailored for remote sensing imagery. For the first time in remote sensing vision--language learning, we propose Multi-Granularity Contrastive Learning (MGCL) and Multi-View Consistency Learning (MVCL) within a unified architecture, explicitly modeling hierarchical semantic correspondence and cross-scale coherence inherent in RS scenes. By jointly aligning global scene structure and localized object-level semantics, GeoAlignCLIP effectively bridges large-scale spatial context and fine-grained discriminative cues.
Specifically,
\textbf{1) }
MGCL consists of Region--Phrase Alignment (RPA) and Hard-Negative Alignment (HNA). RPA aligns region-level visual features with phrase-level textual descriptions, capturing object-aware and structural semantics beyond global CLIP representations. HNA constructs semantically similar yet contextually conflicting hard negatives to address high visual similarity and subtle semantic variations common in RS imagery, thereby enhancing fine-grained discrimination.
\textbf{2) }
MVCL enforces cross-scale coherence within both modalities to handle scale variation and cropping sensitivity in remote sensing. On the visual side, Visual Intra-Consistency (VIC) aligns ROI-view and crop-view representations to mitigate semantic drift. On the textual side, Hierarchical Textual Consistency (HTC) organizes brief and detailed descriptions into a unified hierarchical space, ensuring consistent alignment from global scene semantics to object-level attributes.

To support this learning framework, we construct \textbf{RSFG-100k}, a fine-grained remote sensing dataset consisting of 100k images paired with over 400k hierarchical textual annotations. Each image is annotated with multi-level descriptions, including full-scene captions, region-level statements, and phrase-level labels, together with curated hard negatives. This layered supervision links global scene semantics with localized fine-grained cues in RS imagery.

Extensive experiments demonstrate that GeoAlignCLIP substantially outperforms previous RS vision-language models on fine-grained understanding, open-vocabulary detection, image-text retrieval, and zero-shot classification tasks. Furthermore, GeoAlignCLIP exhibits strong scalability, achieving faster convergence and higher accuracy as the training corpus expands. Our contributions are threefold:

\begin{itemize}
    \item We introduce \textbf{GeoAlignCLIP}, a multi-granularity framework integrating cross-modal fine-grained alignment and multi-view consistency learning for RS imagery.
	\item We construct the \textbf{RSFG-100k}, a fine-grained RS dataset with hierarchical supervision for global-to-local visual-semantic understanding.
	\item We achieve SOTA performance across multiple benchmarks, significantly improving fine-grained recognition, spatial reasoning, and semantic consistency.
\end{itemize}

\vspace{-0.7cm}
\section{Related Works}
\label{sec:2_related_works}
\vspace{-0.2cm}

\noindent\textbf{Vision-Language Models.}
VLMs effectively learn visual and semantic representations by associating images with textual descriptions. CLIP~\cite{radford2021learning}, as a representative model, employs contrastive learning to align image and text features, enabling strong zero-shot generalization across various tasks such as image classification~\cite{sammani2024interpreting}, cross-modal retrieval~\cite{li2024ckdh}~\cite{wang2025cross}, and open-vocabulary detection~\cite{zhu2024survey}.
However, applying VLMs to remote sensing imagery remains challenging due to the unique imaging perspectives, complex backgrounds, and densely distributed objects in this domain. Additionally, the lack of large-scale annotated image-text datasets hinders effective model adaptation. Consequently, despite recent progress, RS VLMs for remote sensing are still in their early stages~\cite{weng2025vision}.

\noindent\textbf{Remote Sensing Vision-Language Models.}
Recent advances in VLMs have greatly promoted multimodal learning in remote sensing, particularly for image–text alignment and cross-modal understanding. Contrastive learning–based methods~\cite{zhou2024visionlanguagegeofoundationmodelsurvey}, such as GeoCLIP~\cite{vivanco2023geoclip}, SkyCLIP~\cite{wang2024skyscript}, and GeoRSCLIP~\cite{zhang2024rs5m}, align remote-sensing images and texts within a shared semantic space, improving retrieval and classification performance. Moreover, models integrating large language models (LLMs)~\cite{brown2020language}\cite{touvron2023llamaopenefficientfoundation}~\cite{liu2023visual}, for example GeoChat~\cite{kuckreja2024geochat}, VHM~\cite{pang2025vhm}, and SkySenseGPT~\cite{luo2024skysensegptfinegrainedinstructiontuning}, combine a CLIP-ViT vision encoder with LLMs through a projection layer, enabling multimodal reasoning via instruction tuning.
Despite these advances, contrastive-based RS VLMs primarily focus on global semantic alignment and often struggle to capture fine-grained spatial relations and object-level details. LLM-based models, while improving interpretability and instruction following, still suffer from limited domain adaptation and weak grounding of visual features in complex remote-sensing scenes. Achieving fine-grained yet semantically consistent alignment therefore remains a central challenge for RS VLMs.

\noindent\textbf{Fine-Grained Alignment.} 
Fine-grained alignment aims to capture subtle differences between images and texts, particularly in terms of object details and spatial relationships, playing a crucial role in improving semantic understanding. To alleviate the performance limitations caused by the lack of fine-grained alignment, recent research has made improvements in several areas, including the capture of fine-grained features~\cite{chen2025lrsclip}\cite{li2025fine}, fine-grained alignment between visual and language features~\cite{zhang2024language}, and the construction of fine-grained image-text data~\cite{zhan2023rsvg}\cite{li2024vrsbench}. These advancements have provided important technical and data foundations for fine-grained alignment in remote sensing images. However, challenges remain, such as insufficient local semantic alignment, the difficulty in balancing global and local features, and the limitations of current datasets. Therefore, there is still significant potential for further development in fine-grained alignment for remote sensing tasks.

\vspace{-0.4cm}
\section{Methodology}
\label{sec:3_method}
\vspace{-0.2cm}

\begin{figure}[t]
    \centering
    \includegraphics[width=0.95\textwidth]{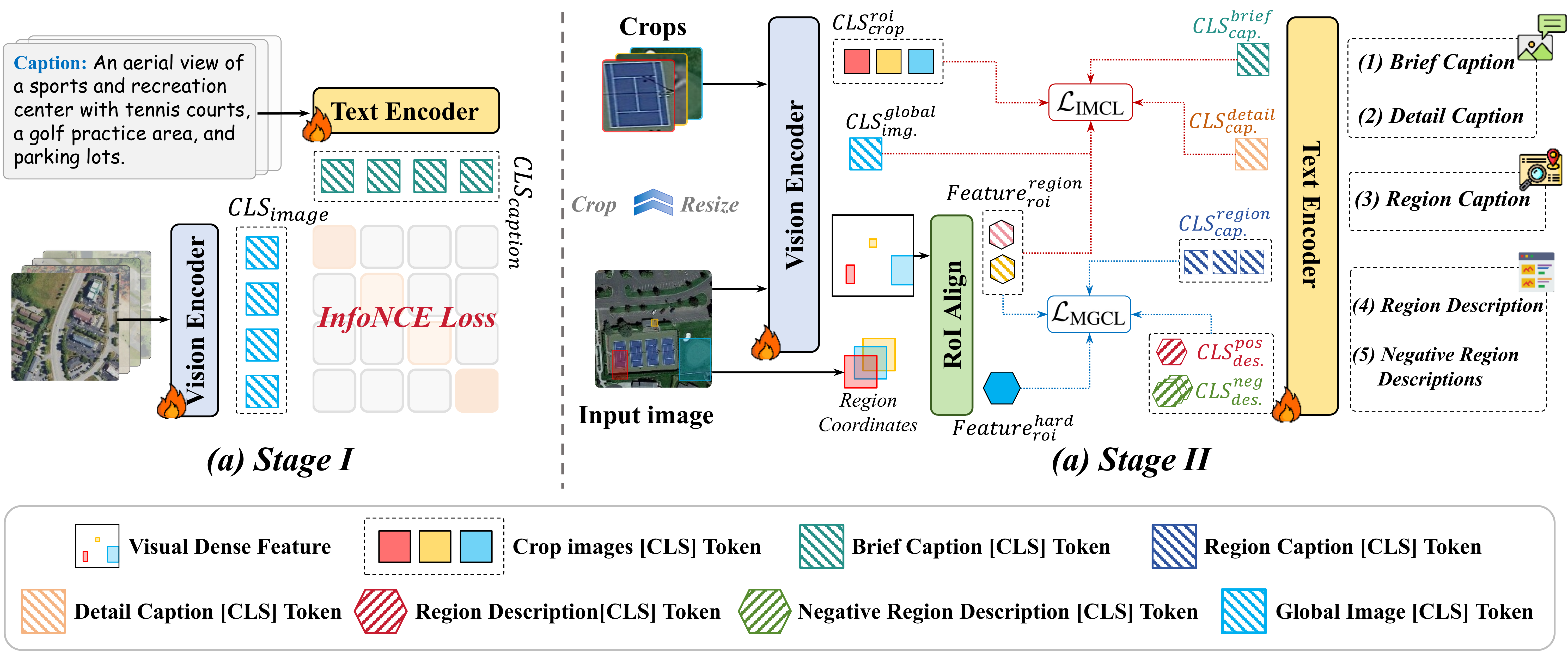}
    \vspace{-0.2cm}
    \caption{
        Overall architecture and training pipeline of \textbf{GeoAlignCLIP}.
        (a) Stage I performs global contrastive learning.
        (b) Stage II conducts Multi-Granularity Contrastive Learning and Multi-View Consistency Learning.
    }
    \label{fig:framework}
    \vspace{-0.5cm}
\end{figure}

As illustrated in Figure~\ref{fig:framework}, GeoAlignCLIP establishes a vision--language alignment architecture following a two-stage learning framework tailored for remote sensing imagery. Due to complex spatial layouts, large scale variations, and high inter-class similarity in RS scenes, direct fine-grained alignment often suffers from unstable optimization and semantic ambiguity.
In the first stage, the model adopts the global image--text contrastive learning paradigm of CLIP~\cite{radford2021learning}, where contrastive learning is performed between the entire image and its corresponding caption to capture large-scale scene semantics and establish a stable cross-modal embedding space.
In the second stage, we perform multi-granularity contrastive learning to explicitly model hierarchical correspondences inherent in RS imagery. This stage integrates region--phrase level fine-grained alignment and semantic hard-negative learning to enhance subtle discrimination among visually similar scenes. Meanwhile, multi-view consistency learning is imposed within both visual and textual modalities to address scale variation and cropping sensitivity, stabilizing multi-scale representations and preserving structural coherence in the semantic space.

\vspace{-0.3cm}
\subsection{Global Contrastive Learning}
\label{subsec:clip}
CLIP adopts a dual-encoder architecture, where an image encoder and a text encoder  project visual and textual inputs into a shared embedding space. 
Both encoders are trained jointly using a symmetric contrastive objective that aligns matched image-text pairs and separates mismatched ones. 
Given a batch of $N$ image-text pairs $\{(I_i, T_i)\}_{i=1}^{N}$, where $I_i$ and $T_i$ denote the visual input and its corresponding brief caption of the $i$-th sample respectively, the optimization objective is defined as:

\vspace{-0.5cm}
\begin{equation}
\begin{aligned}
\mathcal{L}_{\text{g}} = 
-\frac{1}{2N} \sum_{i=1}^{N} \Big[ \,
& \log \frac{\exp(\mathrm{sim}(v_g^i, t_b^i)/\tau)}
{\sum_{j=1}^{N} \exp(\mathrm{sim}(v_g^i, t_b^j)/\tau)} & \hspace{-5mm} + \log \frac{\exp(\mathrm{sim}(t_b^i, v_g^i)/\tau)}
{\sum_{j=1}^{N} \exp(\mathrm{sim}(t_b^i, v_g^j)/\tau)} \ \Big],
\end{aligned}
\label{eq:clip_loss}
\end{equation}
where $\mathrm{sim}(\cdot,\cdot)$ denotes the cosine similarity, and $\tau$ is a learnable temperature parameter. $v_g^i$ denotes the global image [CLS] token representation extracted from the image encoder, 
and $t_b^i$ represents the brief caption [CLS] token representation obtained from the text encoder. 

\vspace{-0.3cm}
\subsection{Multi-Granularity Contrastive Learning}
\label{subsec:mgvsa}
In the CLIP~\cite{radford2021learning} framework, contrastive learning aligns the entire image with its text description (Section~\ref{subsec:clip}). However, this approach focuses on global image-text alignment and fails to capture fine-grained correspondences between local image regions and specific text phrases, which is essential for tasks like remote sensing. To address this, we introduce Region-Phrase Alignment and Hard-Negative Alignment, which refine the alignment between image regions and text phrases.

\noindent\textbf{Region-Phrase Alignment.}
Region-Phrase Alignment aims to establish fine-grained correspondences between visual regions and their associated textual descriptions. 
We first employ RoIAlign~\cite{he2017mask} to extract region-specific visual features 
$\mathrm{Feature}_{\mathrm{roi}}^{(\text{region-}k)}$ 
from the backbone feature maps. 
Average pooling is subsequently applied within each region to derive a set of region-level visual embeddings 
$\{ v_r^{k} \}_{k=1}^{K}$, 
where $K$ denotes the number of valid regions in the batch. 
For each region, we provide a region-level textual description 
$\{ T_{r}^{k} \}_{k=1}^{K}$ 
that semantically characterizes the corresponding localized content. 
These descriptions are processed by the text encoder to obtain textual embeddings 
$\{ t_r^{k} \}_{k=1}^{K}$. 
A contrastive learning objective is subsequently imposed over the region-level visual and textual embeddings to promote discriminative and fine-grained cross-modal alignment. The objective is defined as:

\vspace{-0.5cm}
\begin{equation}
\begin{aligned}
\mathcal{L}_{\text{RPA}} = 
-\frac{1}{2K} \sum_{k=1}^{K} \Big[
& \log 
\frac{
\exp(\mathrm{sim}(v_r^{k}, t_r^{k})/\tau)
}{
\sum_{l=1}^{K}\exp(\mathrm{sim}(v_r^{k}, t_r^{l})/\tau)
} & \hspace{-3.5mm} + 
\log 
\frac{
\exp(\mathrm{sim}(t_r^{k}, v_r^{k})/\tau)
}{
\sum_{l=1}^{K}\exp(\mathrm{sim}(t_r^{k}, v_r^{l})/\tau)
}
\Big].
\end{aligned}
\end{equation}

\noindent\textbf{Hard-Negative Alignment.}  
In the CLIP~\cite{radford2021learning} framework, negative samples are typically obtained through random sampling, where unpaired image-text pairs within a batch are treated as negatives. 
However, such random negatives often exhibit large semantic gaps and provide insufficient discriminative pressure, making it difficult for the model to distinguish semantically similar but mismatched pairs. 

To address this issue, we introduce Hard-Negative Alignment, which enhances region-level discrimination by incorporating hard negatives that maintain high semantic proximity to positive descriptions while differing in fine-grained attributes. For each region-level visual feature $\mathrm{Feature}_{\mathrm{roi}}^{(\text{region-}k)}$ with its embedding $v_r^{k}$, we denote the [CLS] representation of the positive region description as $t_{r}^{k,1}$. We then select $Q-1$ hard negative region--description embeddings $\{ t_{r,j}^{k} \}_{j=2}^{Q}$, which remain close to $t_{r}^{k,1}$ in the semantic space yet introduce controlled attribute-level or other fine-grained variations (see Section~\ref{sec:4_dataset_construction} for details). The objective of Hard-Negative Alignment is defined as:

\begin{equation}
\mathcal{L}_{\text{HNA}} = 
-\frac{1}{K} \sum_{k=1}^{K}
\log
\frac{
\exp(\mathrm{sim}(v_r^{k}, t_{r}^{k,1})/\tau)
}{
\sum_{j=1}^{Q} \exp(\mathrm{sim}(v_r^{k}, t_{r}^{k,j})/\tau)
}.
\end{equation}

\vspace{-0.5cm}
\subsection{Multi-View Consistency Learning}
\label{subsec:imcl}
\vspace{-0.2cm}

While cross-modal alignment bridges visual and textual semantics, stable representation learning also requires consistent visual semantics across different extraction views. In remote sensing imagery, region features derived from feature maps and those obtained by cropping the pixel space may diverge due to scale variations and structural complexity, causing semantic drift. To mitigate it, we introduce multi-view consistency learning that enforces visual intra-modal consistency between these two types of region representations, and further imposes hierarchical textual consistency across phrase-level, region-level, and sentence-level descriptions. This design stabilizes visual semantics across views and strengthens cross-modal alignment.

\noindent\textbf{Visual Intra-Consistency.}  
Objects in remote sensing scenes exhibit considerable variations in scale and spatial configuration.  
Global features extracted from the entire image emphasize scene-level semantics but may overlook local details, while features obtained from cropped regions focus on object-level semantics yet lose global contextual integrity.  
As noted by Cai et al.~\cite{pmlr-v267-cai25d}, cropping in the raw image space can help models attend to local object features but inevitably compromises the overall semantic coherence of the category representation.  

Inspired by this observation, we introduce a visual intra-consistency constraint to maintain stable semantics between local and global visual perspectives. For each region-image pair indexed by $k$, we extract two types of visual features: \textbf{1)} an ROI-view feature $v_r^{k}$ aggregated from the dense global feature map, and \textbf{2)} a cropped-view feature $v_{crop}^{k}$ obtained from the cropped image patch, where the [CLS] token represents the region-level embedding. Both representations originate from the same underlying image and capture complementary contextual and fine-grained information. To enforce semantic coherence across spatial scales, we minimize the cosine distance between each ROI-crop feature pair:

\vspace{-0.2cm}
\begin{equation}
\mathcal{L}_{\text{VIC}} =
\frac{1}{K}\sum_{k=1}^{K}
\Big(1 - \mathrm{sim}(v_r^{k}, v_{crop}^{k})\Big).
\label{eq:vicl_loss}
\end{equation}

\noindent\textbf{Hierarchical Textual Consistency.} 
Finer-grained text descriptions may align more accurately with specific regions of an image but not necessarily with the image as a whole~\cite{li2024visual}.  
Such semantic granularity discrepancy often leads to unstable cross-modal alignment when relying on a single-level textual representation.  
To mitigate this issue, we propose hierarchical textual consistency that enhances alignment from the textual perspective, enabling the model to maintain consistency between global scene semantics and local fine-grained textual details.

Hierarchical textual consistency is constructed using the standard comparison alignment objective applied at two complementary textual granularities. Specifically, we compute a concise-level contrastive loss between the global visual embedding $v_g^i$ and the brief caption $t_b^i$, and a detail-level contrastive loss between $v_g^i$ and the detail caption $t_d^i$, both following the formulation in Eq.~\eqref{eq:clip_loss}. The overall objective of this component is denoted as $\mathcal{L}_{\text{HTC}}$, which jointly enforces consistency between global scene semantics and fine-grained textual details.

\subsection{Overall Objective}
\label{subsec:overall_objective}

Leveraging a two-stage optimization process, GeoAlignCLIP gradually forms a unified, hierarchically aligned vision–language representation space for remote sensing data. The complete learning objective is given by:

\begin{equation}
\hspace{-1mm} \mathcal{L} =
\underbrace{
\lambda_{1}\mathcal{L}_{\text{g}}
}_{\text{Stage I}}
+
\underbrace{
\lambda_{2}\mathcal{L}_{\text{RPA}} +
\lambda_{3}\mathcal{L}_{\text{HNA}}
}_{\text{Stage II: MGCL}}
+
\underbrace{
\lambda_{4}\mathcal{L}_{\text{VIC}} +
\lambda_{5}\mathcal{L}_{\text{HTC}}
}_{\text{Stage II: MVCL}}.
\label{eq:total_loss}
\end{equation}

\section{RSFG-100k Dataset Construction}
\label{sec:4_dataset_construction}

This section introduces the construction process of the proposed RSFG-100k dataset, with additional implementation details available in Appendix. Through multi-source data integration, hierarchical annotation generation, cross-dataset harmonization, and multi-level quality control, RSFG-100k provides semantically consistent, fine-grained, and high-confidence image–text pairs for optical remote sensing understanding. The resulting dataset establishes a unified and extensible foundation for multimodal pretraining and downstream vision–language tasks.

\noindent\textbf{Data Sources.}
RSFG-100k is constructed through the integration of multiple publicly available remote sensing datasets with complementary characteristics, including DE-Dataset~\cite{li2025describeearth}, VRSBench~\cite{li2024vrsbench}, GeoPixelD~\cite{shabbir2025geopixel}, RSVG~\cite{zhan2023rsvg,yang2024mgimm}. These datasets collectively provide diverse spatial resolutions, scene categories, and annotation forms, forming a large-scale basis for high-quality image–text pair construction. RSFG-100k integrates more than 400k samples across three major sources, covering caption-level, region-level, and object-level annotations.

\noindent\textbf{Quality Assessment.} 
We first perform preprocessing and filtering on multi-source remote sensing images to remove low-quality or semantically invalid samples, ensuring the reliability of the input data. This rule-based procedure automatically imposes constraints on image clarity, texture density, and content ratio. Image informativeness is evaluated using brightness and texture variance to discard blurry or near-empty samples, while images with valid region masks exceeding 75\% of the total area are excluded to prevent semantic bias caused by dominant objects or extreme compositions.

\noindent\textbf{Semantic Completion and Enhancement.} To address annotation-level and semantic disparities across different datasets, we construct a unified text generation framework based on Qwen3-VL~\cite{yang2025qwen3}. Specifically, the pipeline consists of four steps: \textbf{1)} generating multi-granularity global captions, where Qwen3-VL produces both brief scene-level summaries and detailed descriptions capturing spatial layouts and contextual relations; \textbf{2)} producing region-level descriptions by incorporating geometric and spatial attributes derived from bounding box dimensions, object centroids, and spatial distributions; \textbf{3)} enriching object-level details using visual cues such as color and texture extracted from local pixel statistics, which are embedded into structured prompts to ensure spatially grounded semantics; and \textbf{4)} synthesizing hard negatives by applying controlled semantic perturbations, such as attribute replacement, orientation inversion, and category-level substitutions.

\noindent\textbf{Harmonization and Quality Review.} 
To ensure data standardization, all samples are uniformly encapsulated into a standard JSON schema encompassing multi-granularity descriptions, hard negatives, and normalized spatial coordinates. To guarantee annotation precision, we implement a dual-tier automated-manual quality control loop. 
\textbf{1) Automated phase.} We first utilize Qwen3-VL~\cite{yang2025qwen3} for an LLM-QA Consistency Check to quantitatively verify image-text alignment. Furthermore, to rigorously validate the integrity of the evaluation splits and preclude potential data leakage, we conduct a dual-modality leakage analysis, as illustrated in Figure~\ref{fig:data_check}. The bivariate density plots for both textual (lexical vs.\ semantic) and visual (structural vs.\ semantic) modalities demonstrate that all test-train pairs densely populate low-similarity regions. Crucially, no sample pairs cross the predefined semantic danger thresholds (red dashed lines), and exact image duplicates are strictly absent (MD5 matches, $N=0$). This absolute isolation of the test set ensures that subsequent evaluations reflect genuine model generalization rather than dataset memorization.
\textbf{2) Manual review phase.} Three domain experts conduct independent cross-validation on a randomly sampled subset of 20k instances (20\% of the total data). As detailed in Table~\ref{tab:audit_summary}, we formulate a targeted Error Taxonomy to categorize annotation noise. The review process yields a high inter-rater agreement (Cohen's Kappa $\kappa = 0.82$) and an overall pass rate of 94.1\%. Ultimately, this rigorous cleansing and verification pipeline ensures high data fidelity, providing a reliable and leakage-free foundational corpus for the subsequent training of VLMs.

\begin{table}[htbp]
  \vspace{-0.6cm}
  \centering
  \begin{minipage}{0.6\textwidth}
    \centering
    \includegraphics[width=\linewidth]{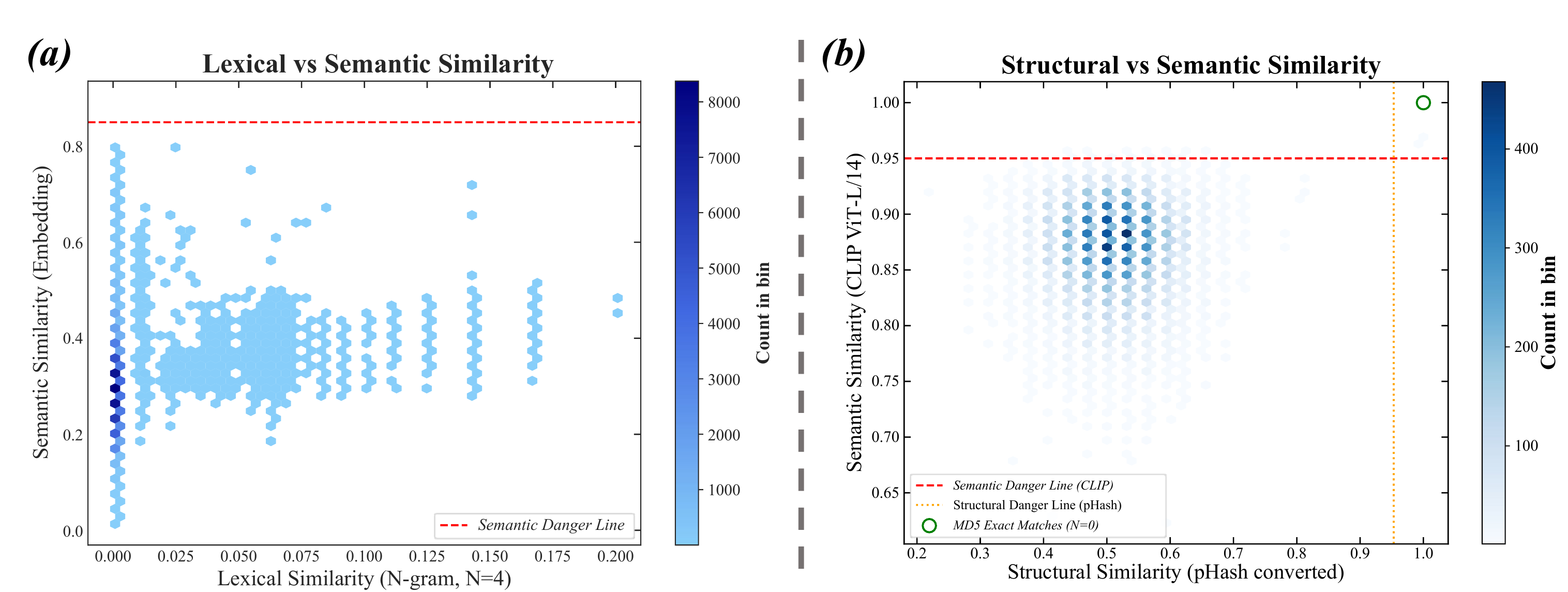}
    \captionof{figure}{Dual-modality data leakage evaluation. (a) Lexical-semantic and (b) structural-visual distribution plots.}
    \label{fig:data_check}
  \end{minipage}
  \hfill 
  \begin{minipage}{0.38\textwidth}
    \centering
    \renewcommand{\arraystretch}{1}
    \resizebox{\linewidth}{!}{
    \begin{tabular}{@{}l|c|c |p{4.5cm} @{}} 
    \toprule
    \textbf{Type} & \textbf{Vol.} & \textbf{Pass} & \textbf{Error Taxonomy} \\ \midrule
    
    Global Cap. & 5k & 92.4 & 1) Subjective Hallucination. 2) Unobservable Inferences. \\ \midrule
    
    Region Desc. & 5k & 95.6 & 1) Metric Ambiguity. 2) Spatial Ambiguity. \\ \midrule
    
    Hard Neg. & 8k & 93.8 & 1) Semantic Distortion. 2) Linguistic Anomalies. \\ \midrule
    
    B.box & 2k & 98.0 & 1) Geometric Offsets. 2) Projection Distortions. \\ \midrule
    
    \textbf{Overall} & \textbf{20k} & \textbf{94.1} & \textbf{Cohen’s Kappa: 0.82}\\ \bottomrule
    \end{tabular}}
    \captionof{table}{Statistical summary of manual review and annotation error taxonomy for RSFG-100k.}
    \label{tab:audit_summary}
  \end{minipage}
\vspace{-1.8cm}
\end{table}

\section{Experiments}
\label{sec:5_experiments}
\vspace{-0.2cm}

This section presents comprehensive experiments to evaluate GeoAlignCLIP on fine-grained understanding, region-level classification, open-vocabulary object detection, and image–text retrieval benchmarks. Additional quantitative and qualitative results are provided in Appendix.

\vspace{-0.3cm}
\subsection{Implementation Details}
GeoAlignCLIP employs ViT-B/16 and ViT-L/14 as visual encoders and applies the Knowledge-Preserved Stretching strategy~\cite{zhang2024long} to extend the text encoder to 248 tokens.
Training is conducted in two stages. 
In the first stage, we pre-train the model on the RSTeller~\cite{ge2025rsteller} dataset for 1 epoch using AdamW with a learning rate of $1\times10^{-4}$ and a weight decay of 0.05. A cosine decay schedule with 200 warm-up iterations is applied. In the second stage, we fine-tune the model on RSFG-100k for 5 epochs using AdamW with a learning rate of $1\times10^{-6}$ and a weight decay of $1\times10^{-3}$, employing 50 warm-up iterations. The per-device batch size is 32. Training is conducted with DeepSpeed ZeRO-2, bf16 mixed precision, and gradient checkpointing.

\vspace{-0.3cm}
\subsection{Comparisons with state-of-the-art Methods}

\begin{table}[t]
    \centering
    \renewcommand{\arraystretch}{0.75}
    \setlength{\tabcolsep}{3pt}
    \resizebox{\textwidth}{!}{
        \begin{tabular}{ll||cc|ccc||cc|cc}
        \toprule
        \multirow{3}{*}{\textbf{Method}} 
        & \multirow{3}{*}{\textbf{Version}} 
        & \multicolumn{5}{c|}{\textbf{Fine-grained Understanding}} 
        & \multicolumn{4}{c}{\textbf{Region-level Classification}} \\
        \cmidrule(lr){3-7} \cmidrule(lr){8-11}
        & & \multicolumn{2}{c|}{\textbf{RRSIS-HR~\cite{liu2025cadformer}}}
        & \multicolumn{2}{c|}{\textbf{CHOICE-ins.~\cite{anchoice}}} 
        & \textbf{CHOICE-img.~\cite{anchoice}}
        & \multicolumn{2}{c|}{\textbf{NWPU-VHR-10~\cite{cheng2014multi}}}
        & \multicolumn{2}{c}{\textbf{RRSIS-D~\cite{liu2024rotated}}} \\
        \cmidrule(lr){3-4} \cmidrule(lr){5-6} \cmidrule(lr){7-7} \cmidrule(lr){8-9} \cmidrule(lr){10-11}
        &  & \textit{Acc@1} & \textit{Acc@5} & \textit{Acc@1} & \textit{Acc@5} & \textit{Acc.} & \textit{Acc@1} & \textit{Acc@5} & \textit{Acc@1} & \textit{Acc@5} \\
        \midrule
        
        \rowcolor{gray!15} \multicolumn{11}{c}{\textit{\textbf{CLIPs based on Natural Images}}} \\
        \cmidrule{1-11}
        \multirow{2}{*}{CLIP~\cite{radford2021learning}} 
             & ViT-B/16 & 19.41 & 60.66 & 50.00 & 91.60 & 72.00 & 12.22 & 54.45 & 41.68 & 73.63 \\
             & ViT-L/14 & 22.18 & 64.47 & 32.80 & 86.20 & 88.00 & 11.87 & 56.90 & 47.47 & 82.65 \\  
        \cmidrule{2-11}
        \multirow{2}{*}{LongCLIP~\cite{zhang2024long}}
             & ViT-B/16 & 13.00 & 58.23 & 27.80 & 90.80 & 82.50 & 56.94 & 83.44 & 40.79 & 69.21 \\
             & ViT-L/14 & 13.52 & 66.72 & 35.80 & 91.60 & 88.00 & 71.64 & 93.62 & 48.09 & 80.10 \\ 
        \cmidrule{2-11}
        \multirow{2}{*}{FG-CLIP~\cite{xie2025fg}}  
             & ViT-B/16 & 27.38 & 73.66 & 13.60 & 87.60 & 87.50 & 86.16 & 98.27 & 53.64 & 84.79 \\
             & ViT-L/14-336 & \underline{32.06} & 71.23 & 18.40 & 89.00 & \underline{89.50} & 88.37 & \underline{99.86} & 63.00 & 93.26 \\ 
             
        \midrule
        \rowcolor{gray!15} \multicolumn{11}{c}{\textit{\textbf{CLIPs based on Remote Sensing Images}}} \\
        \cmidrule{1-11}
        \multirow{2}{*}{RemoteCLIP~\cite{liu2024remoteclip}} 
             & ViT-B/32 & 6.76 & 31.89 & 61.20 & 97.60 & 76.50 & 66.01 & 91.41 & 42.60 & 64.48 \\
             & ViT-L/14 & 8.32 & 36.40 & 61.40 & \textbf{98.80} & 77.50 & 15.25 & 42.79 & 51.76 & 72.26 \\ 
        \cmidrule{2-11}
        \multirow{3}{*}{SkyCLIP~\cite{wang2024skyscript}}  
             & ViT-B/32 & 16.46 & 50.09 & 12.40 & 84.60 & 78.50 & 8.32 & 67.18 & 1.66 & 17.83 \\
             & ViT-L/14-pct30 & 6.41 & 40.03 & 49.80 & 92.40 & 85.00 & 81.82 & 98.41 & 53.04 & 87.45 \\
             & ViT-L/14-pct50 & 2.77 & 30.16 & 48.20 & 91.40 & 84.50 & 83.64 & 98.76 & 54.04 & 87.91 \\ 
        \cmidrule{2-11}
        \multirow{3}{*}{GeoRSCLIP~\cite{zhang2024rs5m}} 
             & ViT-B/32 & 11.09 & 49.05 & 4.00 & 58.60 & 83.00 & 73.15 & 98.55 & 52.07 & 79.51 \\
             & ViT-L/14 & 11.09 & 51.47 & 2.60 & 58.20 & 83.50 & 74.26 & 95.62 & 58.27 & 82.41 \\
             & ViT-L/14-336 & 11.44 & 58.23 & 4.40 & 61.60 & 83.50 & 72.77 & 91.55 & 57.64 & 81.98 \\ 
        \cmidrule{2-11}
        \multirow{2}{*}{LRSCLIP~\cite{chen2025lrsclip}}                 
             & ViT-B/16 & 27.04 & \underline{78.16} & 2.80 & 59.00 & 86.50 & 64.80 & 95.96 & 45.48 & 80.99 \\
             & ViT-L/14 & 26.69 & 76.78 & 20.20 & 86.00 & 20.20 & 67.60 & 94.93 & 53.28 & 85.38 \\ 
             
        \midrule
        \rowcolor{gray!15} \multicolumn{11}{c}{\textit{\textbf{Remote Sensing Large Vision Language Models}}} \\
        \cmidrule{1-11}
        GeoChat~\cite{kuckreja2024geochat} 
             & Vicuna v1.5-7B & 22.01 & - & 43.20 & - & 16.50 & 49.54 & - & 54.71 & - \\
        SkysenseGPT~\cite{luo2024skysensegptfinegrainedinstructiontuning} 
             & Vicuna v1.5-7B & 21.32 & - & 41.60 & - & 16.50 & 74.31 & - & 73.44 & - \\
        VHM~\cite{pang2025vhm} 
             & Vicuna-7B & 24.26 & - & \textbf{63.80} & - & 74.00 & 75.69 & - & \underline{77.86} & - \\
        \midrule
        \multirow{2}{*}{\textbf{Ours}}
             & ViT-B/16 & 29.46 & 75.74 & \underline{62.00} & \underline{98.60} & 88.00 & \underline{90.27} & 99.41 & 77.04 & \underline{93.57} \\
             & ViT-L/14 & \textbf{33.45} & \textbf{81.28} & \underline{62.00} & \textbf{98.80} & \textbf{92.00} & \textbf{93.75} & \textbf{99.97} & \textbf{82.89} & \textbf{98.03} \\
        \bottomrule
        \end{tabular}
    }
    \caption{Experimental results on fine-grained understanding and region-level Classification tasks. Acc@1 and Acc@5 denote Top-1 and Top-5 accuracy, respectively.}
    \label{tab:merged_results}
    \vspace{-1.1cm}
\end{table}

\noindent\textbf{Fine-grained Understanding.}
To evaluate the fine-grained perception and vision-language alignment capability of GeoAlignCLIP, we conduct experiments on two challenging benchmarks: RRSIS-HR~\cite{liu2025cadformer} and CHOICE~\cite{anchoice}, where the latter is further divided into region-level (CHOICE-ins.) and image-level (CHOICE-img.) fine-grained tasks. In addition to their inherent complexity, we incorporate textual hard negatives by minimally perturbing key attributes in the ground-truth descriptions, creating semantically shifted yet lexically similar variants. This setup provides a stricter evaluation of the model’s sensitivity to fine-grained attributes and resilience to subtle linguistic ambiguities.

As shown in Table~\ref{tab:merged_results}, general-domain foundational models (e.g., CLIP, LongCLIP) exhibit limited fine-grained recognition on RS benchmarks. While domain-adapted architectures like LRSCLIP yield competitive Top-5 accuracies, and LVLMs (e.g., VHM) achieve notable instance-level metrics (63.80\% Acc@1 on CHOICE-ins.), they generally struggle with comprehensive fine-grained alignment. Conversely, GeoAlignCLIP establishes a new state-of-the-art, securing 33.45\%/81.28\% (Acc@1/Acc@5) on RRSIS-HR and dominating the CHOICE benchmark (92.00\% on CHOICE-img.\ and 62.00\%/98.80\% on CHOICE-ins.), demonstrating a superior capacity to decode precise region-level semantics. Notably, although general-domain fine-grained methods leverage localized alignment to outperform several RS-specific models, their absence of domain pretraining restricts their efficacy in complex scenes. These findings confirm that explicit fine-grained semantic modeling is fundamentally more critical for dense RS imagery than mere domain adaptation.

\noindent\textbf{Region-level Classification.}
To evaluate the model's capability in local semantic recognition, we conduct the Region-level Classification experiment on the NWPU-VHR-10~\cite{cheng2014multi} and RRSIS-D~\cite{liu2024rotated} datasets. Under the zero-shot setting, each annotated bounding box is classified using the model’s prediction, and the performance is quantified by Top-1 and Top-5 accuracy.

As shown in Table~\ref{tab:merged_results}, our method achieves the best performance on both datasets, reaching 93.75\% / 99.97\% (Acc@1 / Acc@5) on NWPU-VHR-10 and 82.89\% / 98.03\% on RRSIS-D. It is noteworthy that FG-CLIP, although trained solely on natural images, achieves comparable or even superior results to some RS-specific CLIPs, highlighting the importance of fine-grained visual modeling for object-level recognition. Overall, models trained on remote sensing data consistently outperform general CLIP baselines, demonstrating the effectiveness of domain-specific data in enhancing adaptation to remote sensing imagery.
In summary, these results indicate that our method can accurately recognize the category of each target region under zero-shot conditions, reflecting significant improvements in local semantic alignment and fine-grained understanding.

\begin{figure}[t]
    \centering
    \begin{minipage}[c]{0.55\linewidth}
        \centering
        \scriptsize
        \renewcommand{\arraystretch}{0.85}
        \resizebox{\linewidth}{!}{
        \begin{tabular}{c|l|c|>{\color{gray!100}}c>{\color{gray!100}}c}
            \toprule
            \textbf{Dataset} & \textbf{Method} & \textit{mAP\(_{\text{n}}\)} & \textit{mAP\(_{\text{b}}\)} & \textit{mAP} \\
            \midrule
            \multirow{7}{*}{DIOR~\cite{li2020object}}
                & ViLD~\cite{guopen} & 0.70 & 47.70 & 39.30 \\
                & GLIP~\cite{li2022grounded} & 5.50 & 47.90 & 39.40 \\
                & GroundingDINO~\cite{liu2024grounding} & 9.80 & 48.70 & 39.50 \\
                & FG-CLIP~\cite{xie2025fg} & 10.90 & \underline{64.90} & 54.10  \\
                & LRSCLIP~\cite{chen2025lrsclip} & \underline{15.20} & 64.50 & \underline{54.70}  \\
                & GeoRSCLIP~\cite{zhang2024rs5m} & 14.00 & 64.40 & 54.30  \\
                & \cellcolor{gray!15}\textbf{Ours} & \cellcolor{gray!15}\textbf{17.10} & \cellcolor{gray!15}\textbf{65.00} & \cellcolor{gray!15}\textbf{55.40}   \\
            \midrule
            \multirow{7}{*}{DOTAv1.0~\cite{Xia_2018_CVPR}}
                & ViLD~\cite{guopen} & 2.50 & \textbf{67.70} & \textbf{59.00} \\
                & GLIP~\cite{li2022grounded} & 8.00 & \underline{65.20} & 57.50 \\
                & GroundingDINO~\cite{liu2024grounding} & 1.90 & 61.00 & 53.10 \\
                & FG-CLIP~\cite{xie2025fg} & \underline{20.70} & 63.70 & 57.90 \\
                & LRSCLIP~\cite{chen2025lrsclip} & 19.40 & 61.00 & 55.40 \\
                & GeoRSCLIP~\cite{zhang2024rs5m} & 9.00 & 59.90 & 53.10 \\ 
                & \cellcolor{gray!15}\textbf{Ours} & \cellcolor{gray!15}\textbf{25.50} & \cellcolor{gray!15}63.30 & \cellcolor{gray!15}58.30  \\
            \bottomrule
        \end{tabular}
        }
        \vspace{-0.3cm}
        \captionof{table}{Comparisons of open-vocabulary object detection task on the DIOR and DOTAv1.0. \textit{mAP\(_{\text{n}}\)} and \textit{mAP\(_{\text{b}}\)} denote mean average precision on novel and base classes, respectively.}
        \label{tab:region_open}
    \end{minipage}
    \hfill
    \begin{minipage}[c]{0.335\linewidth}
        \centering
        \includegraphics[width=\linewidth]{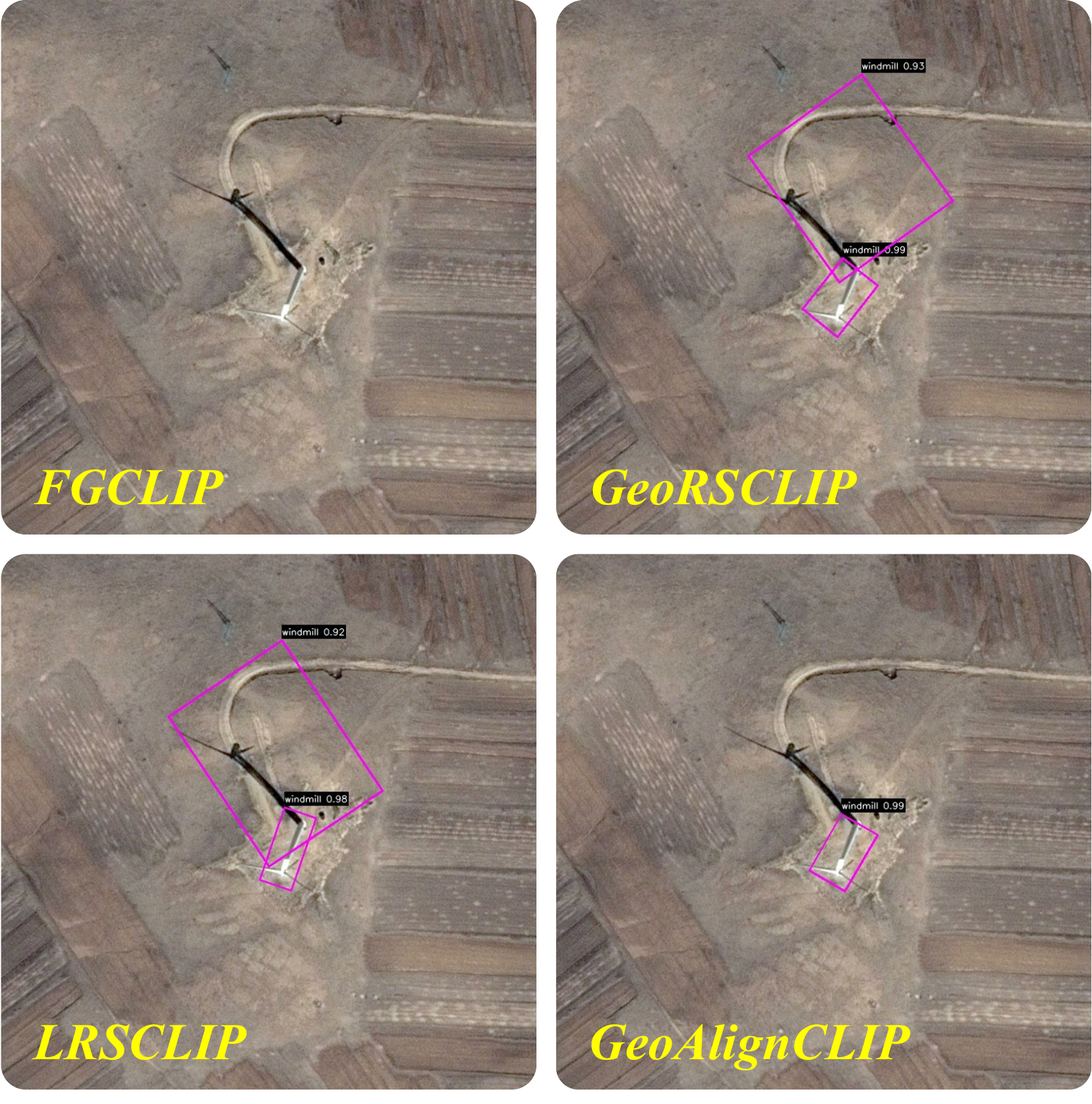}  
        \vspace{-0.65cm}
        \captionof{figure}{Visualization results of open-vocabulary object detection.}
        \label{fig:ovd_vis}
    \end{minipage}
\vspace{-0.8cm}
\end{figure}

\noindent\textbf{Open-Vocabulary Object Detection.}
We evaluate the open-vocabulary detection (OVD) capability of GeoAlignCLIP by adopting the CastDet framework~\cite{li2024toward}, where different CLIP variants are directly substituted into the architecture to benchmark their cross-domain generalization. Adhering to the standard base-to-novel transfer protocol, the detector is trained exclusively on the base-category annotations of DIOR~\cite{li2020object} and DOTAv1.0~\cite{Xia_2018_CVPR}. Both training and evaluation are conducted on strictly isolated official splits to preserve benchmark integrity. While all training images contribute to the pseudo-labeling pipeline, the final performance is reported solely on the official test sets, focusing primarily on novel-class precision ($mAP_{n}$). To guarantee a rigorous and fair comparison, all experimental configurations and hyperparameters remain strictly identical across the evaluated CLIP models, as detailed in Table~\ref{tab:training-params}.

\begin{table}[h]
  \vspace{-0.5cm}
  \centering
  \resizebox{1\textwidth}{!}{
  \begin{tabular}{ll|ll|ll}
    \toprule
    \textbf{Parameter} & \textbf{Value} & \textbf{Parameter} & \textbf{Value} & \textbf{Parameter} & \textbf{Value} \\
    \midrule
    Backbone Network & ResNet-50 & Pseudo-Label Threshold & 0.5 & Max Training Iterations & 10,000 \\
    Input Image Size & 1024 $\times$ 1024 & RPN Filtering Threshold & 0.9 & Learning Rate (LR) & 0.01 \\
    Supervised Loss Weight & 1.0 & Dynamic Queue Start & 2,000 & LR Warm-up Iterations & 500 \\
    Unsupervised Loss Weight & 2.0 & Neck Out Channels & 256 & Momentum & 0.9 \\
    FC Out Channels & 768 & RPN In Channels & 256 & Weight Decay & 0.0001 \\
    BBox Head In Channels & 256 & Pad Size Divisor & 32 & Image Format & RGB \\
    \bottomrule
  \end{tabular}
  }
  \caption{Experimental configurations for open-vocabulary detection.}
  \label{tab:training-params}
\end{table}

Table~\ref{tab:region_open} presents the results. On DIOR, natural-image open-vocabulary detectors such as ViLD~\cite{guopen}, GLIP~\cite{li2022grounded}, and GroundingDINO~\cite{liu2024grounding} exhibit limited transferability to remote-sensing data, with mAP\(_{\text{n}}\) scores of 0.7\%, 5.5\%, and 9.8\%, respectively. RS-specific CLIPs, including LRSCLIP and GeoRSCLIP, improve performance, reaching 15.2\% and 14.0\%. Our method further enhances the state of the art to 17.1\% mAP\(_{\text{n}}\), demonstrating superior cross-domain generalization. A similar trend is observed on DOTAv1.0, where our method outperforms FG-CLIP~\cite{xie2025fg} by 4.8\%, achieving 25.5\% mAP on novel classes, without compromising performance on base categories. Qualitative results, shown in Figure~\ref{fig:ovd_vis}, further highlight the advantages of our approach. Methods like FGCLIP, GeoRSCLIP, and LRSCLIP fail to detect the wind turbine or misidentify its shadow as the turbine. In contrast, GeoAlignCLIP successfully detects the turbine, demonstrating its robustness in handling challenging remote-sensing scenarios. These results confirm the effectiveness of our alignment strategy in improving both detection accuracy and generalization to unseen categories.

\begin{table}[t]

    \begin{center}
    \resizebox{\textwidth}{!}{
    \begin{tabular}{ll|ccccccccccccccccccc}
        \toprule
        \multirow{3}{*}{\textbf{Method}} & \multirow{3}{*}{\textbf{Backbone}} 
        & \multicolumn{6}{c}{\textbf{RSICD~\cite{lu2017exploring}}} 
        & \multicolumn{6}{c}{\textbf{RSITMD~\cite{yuan2021exploring}}}  
        & \multicolumn{6}{c}{\textbf{UCM-Caption~\cite{lu2017exploring}}} \\
        \cmidrule(lr){3-8} \cmidrule(lr){9-14} \cmidrule(lr){15-20}
        & & \multicolumn{3}{c}{I2T} & \multicolumn{3}{c}{T2I}    
        & \multicolumn{3}{c}{I2T} & \multicolumn{3}{c}{T2I}    
        & \multicolumn{3}{c}{I2T} & \multicolumn{3}{c}{T2I} \\
        \cmidrule(lr){3-5} \cmidrule(lr){6-8} 
        \cmidrule(lr){9-11} \cmidrule(lr){12-14} 
        \cmidrule(lr){15-17} \cmidrule(lr){18-20}
        & & \textit{R@1} & \textit{R@5} & \textit{R@10} & \textit{R@1} & \textit{R@5} & \textit{R@10}   
        & \textit{R@1} & \textit{R@5} & \textit{R@10} & \textit{R@1} & \textit{R@5} & \textit{R@10}   
        & \textit{R@1} & \textit{R@5} & \textit{R@10} & \textit{R@1} & \textit{R@5} & \textit{R@10} \\
        \midrule
        \rowcolor{gray!15} \multicolumn{20}{c}{\textit{ \textbf{CLIPs based on Natural Images}}} \\
        \cmidrule{1-20}
        \multirow{2}{*}{CLIP \cite{radford2021learning}} & ViT-B/16 &5.67  &14.82  &22.6  &5.29  &16.96  &26.77  &7.96  &21.46  &31.42  &8.36  &26.11  &40.84  &8.57  &31.9  &55.71  &9.52  &36.0  &64.1  \\
             & ViT-L/14 &6.59  &17.75  &28.09  &4.96  &18.72  &29.66  &10.62  &27.43  &36.73  &10.04  &31.33  &46.15  &12.38  &41.43  &70.95  &10.57  &43.81  &71.24  \\ \cmidrule{2-20}
        \multirow{2}{*}{LongCLIP \cite{zhang2024long}} & ViT-B/16 &7.69  &20.13  &29.19  &6.68  &20.68  &32.17  &8.85  &23.67  &32.96  &8.72  &33.41  &49.07  &12.86  &41.9  &65.24  &11.81  &45.81  &76.19  \\
        & ViT-L/14 &8.78  &20.4  &28.82  &7.14  &23.48  &35.55  &11.06  &27.43  &38.27  &12.74  &36.9  &53.14  &13.33  &41.9  &67.14  &12.95  &49.62  &81.05  \\ \cmidrule{2-20}
        \multirow{2}{*}{FGCLIP \cite{xie2025fg}} & ViT-B/16 & 9.42 & 21.41 & 30.56 & 8.60 & 25.36 & 37.62 & 13.72 & 28.76 & 36.95 & 14.60 & 38.32 & 52.92 & 11.90 & 38.10 & 65.71 & 12.57 & 47.24 & 75.90 \\
               & ViT-L/14-336 & 11.25 & 25.34 & 34.95 & \underline{11.18} & \underline{29.83} & 43.31 & 16.15 & 33.41 & 45.13 & 15.84 & \underline{42.30} & \textbf{58.10} & 17.62 & 48.57 & 69.05 & 15.43 & \underline{55.62} & 84.00 \\
        \midrule
        \rowcolor{gray!15} \multicolumn{20}{c}{\textit{ \textbf{CLIPs based on Remote Sensing Images}}} \\
        \cmidrule{1-20}
        \multirow{3}{*}{SkyCLIP \cite{wang2024skyscript}} & ViT-B/32 & 6.22 & 17.20 & 26.53 & 6.99 & 21.65 & 33.65 & 9.29 & 24.12 & 34.51 & 10.31 & 33.63 & 49.60 & 15.24 & 40.95 & 63.81 & 11.90 & 46.86 & 77.33 \\
                & ViT-L/14-pct30 & 5.86 & 18.39 & 27.08 & 6.40 & 20.15 & 32.86 & 12.83 & 28.32 & 39.82 & 11.19 & 34.87 & 50.40 & 15.24 & 44.76 & 66.67 & 12.48 & 47.24 & 76.76 \\
                & ViT-L/14-pct50 & 5.86 & 18.30 & 27.08 & 6.02 & 20.31 & 32.00 & 12.17 & 28.32 & 38.50 & 10.75 & 34.69 & 50.58 & 14.76 & 44.76 & 68.10 & 12.19 & 46.86 & 76.29 \\ \cmidrule{2-20}
        \multirow{3}{*}{GeoRSCLIP \cite{zhang2024rs5m}} & ViT-B/32 & 11.25 & 26.53 & 37.15 & 8.45 & 25.01 & 38.35 & 15.49 & 32.96 & 43.58 & 12.35 & 38.58 & 55.31 & 16.67 & 44.76 & 72.38 & 14.76 & 51.05 & 81.81 \\
                & ViT-L/14 & 12.81 & 28.91 & \underline{41.17} & 9.92 & 28.01 & 42.09 & 19.69 & 38.50 & 48.89 & \underline{17.21} & 41.37 & 56.33 & 18.57 & 46.19 & 71.90 & \underline{16.10} & 54.48 & \textbf{86.19} \\
                & ViT-L/14-336 & 12.08 & 27.26 & 39.71 & 10.36 & 28.44 & 42.09 & 20.13 & 37.17 & 48.23 & 16.28 & 42.08 & 57.39 & \underline{19.05} & 49.05 & 70.95 & 15.43 & 54.57 & 84.67 \\ \cmidrule{2-20}
        \multirow{2}{*}{LRSCLIP \cite{chen2025lrsclip}} & ViT-B/16 & 13.08 & 29.83 & 39.52 & 9.15 & 27.43 & 39.65 & \underline{20.58} & \textbf{40.71} & \underline{50.66} & 14.65 & 37.21 & 51.99 & \textbf{20.48} & \textbf{52.86} & \textbf{82.38} & 14.29 & 53.24 & 85.62 \\
                & ViT-L/14 & \textbf{15.65} & \textbf{31.38} & \textbf{41.26} & 10.01 & 28.60 & \underline{43.73} & 20.35 & \underline{39.82} & \textbf{51.77} & \textbf{17.39} & 41.59 & 56.77 & \underline{19.05} & \textbf{52.86} & 77.14 & 15.71 & \textbf{55.81} & 85.24 \\
        \midrule
        \multirow{2}{*}{\textbf{Ours} }
                & ViT-B/16 & 14.00 & 27.81 & 38.06 & 9.55 & 28.07 & 41.92 & 18.36 & 34.29 & 48.67 & 15.00 & 40.49 & 55.49 & 15.71 & \underline{49.52} & 74.29 & 13.81 & 50.57 & 79.43 \\
                & ViT-L/14 & \underline{14.82} & \underline{30.38} & 40.71 & \textbf{11.40} & \textbf{31.29} & \textbf{44.34} & \textbf{21.02} & 37.17 & 49.34 & \textbf{17.39} & \textbf{42.39} & \underline{58.05} & \textbf{20.48} & \textbf{52.86} & \underline{77.62} & \textbf{17.71} & 55.33 & \underline{86.10} \\
        \bottomrule
    \end{tabular}
    }
    \caption{Comparisons of image-text retrieval task on the RSICD, RSITMD, and UCM-Caption. I2T and T2I denote image-to-text retrieval and text-to-image retrieval, respectively. \textit{R@k} represent Recall at top-\textit{k}.}
    \label{sl_it}
\end{center}
\vspace{-1.3cm}
\end{table}

\noindent\textbf{Image-Text Retrieval.}
As reported in Table~\ref{sl_it}, GeoAlignCLIP demonstrates competitive cross-modal retrieval capabilities across the RSICD~\cite{lu2017exploring}, RSITMD~\cite{yuan2021exploring}, and UCM-Caption~\cite{lu2017exploring} benchmarks. Specifically, GeoAlignCLIP achieves the highest text-to-image (T2I) R@1 scores on RSICD (11.40\%) and UCM-Caption (17.71\%), while establishing the leading image-to-text (I2T) R@1 on RSITMD (21.02\%). On the remaining metrics, the model either matches the best baseline results, notably yielding a 20.48\% I2T R@1 on UCM-Caption, or maintains comparable performance against leading domain-adapted architectures like LRSCLIP. Furthermore, GeoAlignCLIP yields substantial improvements over general-domain methods; for instance, its 17.39\% T2I R@1 on RSITMD clearly surpasses both generic CLIP (10.04\%) and LongCLIP (12.74\%). These quantitative results confirm that GeoAlignCLIP effectively enhances local discriminability without compromising the global feature alignment necessary for balanced retrieval tasks.

\subsection{Ablation Study}
We evaluate the individual contributions of the proposed MGCL and MVCL mechanisms in Table \ref{tab:ablation_stage}. Starting from the Stage I baseline that relies solely on global alignment, introducing the Region-Phrase Alignment ($\mathcal{L}_{RPA}$) yields substantial immediate gains, notably elevating the Acc@1 on RRSIS-HR from 25.48\% to 33.28\% and validating the necessity of establishing explicit fine-grained correspondences. Subsequently, incorporating Visual Intra-Consistency ($\mathcal{L}_{VIC}$) significantly improves the $mAP_{n}$ on DOTAv1.0 from 14.80 to 20.20, confirming its efficacy in mitigating semantic drift caused by inherent scale variations in RS imagery by aligning global feature maps with localized ROI crops. The addition of Hard-Negative Alignment ($\mathcal{L}_{NHA}$) further enhances fine-grained discriminability among semantically analogous categories, while the comprehensive integration of Hierarchical Textual Consistency ($\mathcal{L}_{HTC}$) culminates in optimal performance across all benchmarks. Specifically, the unified framework peaks at an Acc@1 of 42.28\% on RRSIS-HR and a 25.50\% $mAP_{n}$ on DOTAv1.0. Overall, these consistent improvements indicate that the synergy between multi-scale spatial consistency and fine-grained semantic mining significantly enhances the robustness of remote sensing vision-language understanding.

\begin{table}[t]
\centering
\setlength{\tabcolsep}{2.2pt}
\renewcommand{\arraystretch}{0.82}
\setlength{\textfloatsep}{6pt}
\setlength{\floatsep}{4pt}
\setlength{\intextsep}{6pt}
\resizebox{0.8\linewidth}{!}{
  \begin{tabular}{cccc|ccccc}
\toprule
\multicolumn{4}{c}{\textbf{Loss}}|
& \multicolumn{1}{c}{\textbf{RRSIS-HR}}
& \multicolumn{1}{c}{\textbf{DOTAv1.0}}
& \multicolumn{2}{c}{\textbf{UCM-Caption}}
& \multicolumn{1}{c}{\textbf{SIRI-WHU}} \\
\cmidrule(lr){1-4}\cmidrule(lr){5-5}\cmidrule(lr){6-6}\cmidrule(lr){7-8}\cmidrule(lr){9-9}

\multicolumn{1}{c}{$\mathcal{L}_{\text{RPA}}$}
& \multicolumn{1}{c}{$\mathcal{L}_{\text{VIC}}$}
& \multicolumn{1}{c}{$\mathcal{L}_{\text{NHA}}$}
& \multicolumn{1}{c}{$\mathcal{L}_{\text{HTC}}$}
& \multicolumn{1}{c}{\textit{Acc@1}}
& \multicolumn{1}{c}{\textit{mAP\(_{\text{n}}\)}}
& \multicolumn{1}{c}{I2T \textit{R@1}}
& \multicolumn{1}{c}{T2I \textit{R@1}}
& \multicolumn{1}{c}{\textit{Acc@1}}\\
\midrule
- & - & - & - & 25.48 & 14.50 & 16.67 & 12.90 & 51.92 \\
$\checkmark$ & - & - & - & 33.28 & 14.80 & 20.00 & 16.57 & 59.21 \\
$\checkmark$ & $\checkmark$ & - & - & 34.66 & 20.20 & 20.00 & 16.67 & 61.96 \\
$\checkmark$ & $\checkmark$ & $\checkmark$ & - & 35.01 & 21.50 & 18.10 & 16.48 & 63.04 \\
$\checkmark$ & $\checkmark$ & - & $\checkmark$ & 35.36 & 16.50 & 20.14 & 16.19 & 63.42 \\
\rowcolor{gray!15}$\checkmark$ & $\checkmark$ & $\checkmark$ & $\checkmark$
& \textbf{42.28} & \textbf{25.50} & \textbf{20.48} & \textbf{17.71} & \textbf{64.21} \\

\bottomrule
\end{tabular}}
\caption{Ablation study of different tasks on the VRSBench, DOTAv1.0, UCM-Caption, and SIRI-WHU datasets.}
\label{tab:ablation_stage}
\vspace{-1cm}
\end{table}

\subsection{Parameter Size and Efficiency}

To assess the computational cost of GeoAlignCLIP, we compare its model parameters (Param. in Millions), per-token inference latency (Lat. in milliseconds), floating-point operations (FLOPs in TeraFLOPs), and retrieval accuracy against baseline methods. As detailed in Tab.~\ref{tab:efficiency_comparison}, we analyze the computational footprint of GeoAlignCLIP. Despite the integration of multi-granular alignment modules, the proposed architecture incurs a marginal parameter overhead of only 1.30M (from 427.62M to 428.92M). While region-specific feature extraction inherently increases theoretical FLOPs (13.65T vs. 5.40T), this additional computational demand primarily comprises highly parallelizable operations (\eg, RoIAlign). On modern high-bandwidth hardware, these localized computations effectively saturate parallel compute cores without extending the sequential depth of the underlying Transformer backbone. 

\begin{wraptable}{r}{0.45\textwidth} 
\vspace{-0.5cm} 
\renewcommand{\arraystretch}{1}
\setlength{\tabcolsep}{3.5pt} 
\centering
\resizebox{\linewidth}{!}{ 
\begin{tabular}{l|cccc}
\toprule
\textbf{Model} & \textit{Param.} & \textit{Lat.} & \textit{FLOPs} & \textit{Acc.} \\
 & \scriptsize{(M)} & \scriptsize{(ms/token)} & \scriptsize{(T)} & \scriptsize{(\%)} \\
\midrule
RemoteCLIP    & 427.62 & 0.1218 & 5.40 & 8.32 \\
SkyCLIP       & 427.62 & 0.1216 & 5.40 & 6.41 \\
GeoRSCLIP     & 427.62 & 0.1215 & 5.40 & 11.44 \\
LRSCLIP       & 427.62 & 0.1215 & 5.40 & 26.69 \\
\rowcolor{gray!15}\textbf{Ours} & 428.92 & 0.1327 & 13.65 & \textbf{33.45} \\
\bottomrule
\end{tabular}
}
\caption{Efficiency comparison. Latency is the per-token inference time measured.}
\label{tab:efficiency_comparison}
\vspace{-10pt} 
\end{wraptable}
Consequently, GeoAlignCLIP achieves a highly competitive per-token inference latency of 0.1327~ms, performing on par with standard CLIP baselines (\eg, 0.1215~ms for LRSCLIP). These metrics substantiate that our significant performance gains---such as the absolute 6.76\% improvement in Acc@1 (from 26.69\% to 33.45\%)---stem from a principled cross-modal architectural design rather than naive capacity scaling, thereby guaranteeing practical deployability for fine-grained remote sensing tasks.

\vspace{-0.4cm}
\subsection{Visualization Study}
\vspace{-0.1cm}

\begin{figure}[t]
    \centering
    \includegraphics[width=0.8\linewidth]{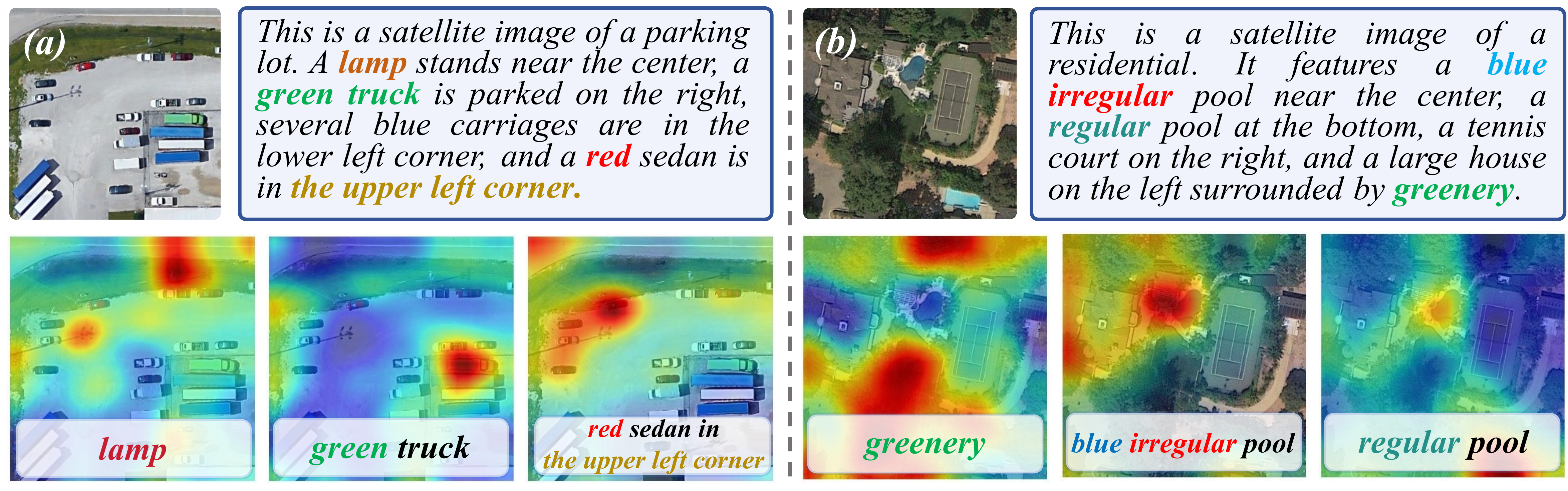}
    \vspace{-0.3cm}
    \caption{
    Visualization of fine-grained vision-language alignment by GeoAlignCLIP. 
    }
    \label{fig:vis_fine}
    \vspace{-0.5cm}
\end{figure}

Figure~\ref{fig:vis_fine} presents qualitative examples illustrating GeoAlignCLIP’s fine-grained vision-language alignment. 
In Figure~\ref{fig:vis_fine}(a), the model accurately localizes multiple region–phrase pairs within a parking lot scene, including the \textit{lamp}, the \textit{green truck}, and the \textit{red sedan in the upper left corner}. 
Each attention map highlights the corresponding region with high spatial precision, indicating that GeoAlignCLIP effectively distinguishes between semantically similar objects under dense layouts and subtle attribute variations. 
Similarly, in the residential scene shown in Figure~\ref{fig:vis_fine}(b), the model successfully aligns descriptive phrases such as \textit{greenery}, \textit{blue irregular pool}, and \textit{the regular pool}, capturing both local textures and global spatial relations. 
Notably, the attention responses reflect clear spatial disentanglement among overlapping entities (e.g., pools and surrounding vegetation), demonstrating the model’s capacity to maintain fine-grained consistency across heterogeneous visual contexts. 
Overall, these results validate that GeoAlignCLIP can accurately ground textual attributes to their visual counterparts, achieving coherent multi-granularity alignment in complex remote sensing imagery.

\vspace{-0.2cm}
\section{Conclusion}
\label{sec:6_conclusion}

In this paper, we presented GeoAlignCLIP, a unified vision-language framework that achieves fine-grained alignment for remote sensing imagery through multi-granularity consistency learning. By integrating region-phrase alignment, semantic hard-negative learning, and multi-view consistency constraints, GeoAlignCLIP effectively balances global and local semantics while enhancing fine-grained discriminability. Extensive experiments on multiple benchmarks demonstrate its superiority across retrieval, classification, and open-vocabulary detection tasks. In future work, we plan to extend our framework to multimodal large vision-language models and explore its scalability to broader geospatial understanding scenarios.

\bibliographystyle{splncs04}
\bibliography{main}

\end{document}